# IMPROVING NERF WITH HEIGHT DATA FOR UTILIZATION OF GIS DATA

*Hinata Aoki and Takao Yamanaka*

Department of Information and Communication Sciences, Sophia University, Japan

## ABSTRACT

Neural Radiance Fields (NeRF) has been applied to various tasks related to representations of 3D scenes. Most studies based on NeRF have focused on a small object, while a few studies have tried to reconstruct large-scale scenes although these methods tend to require large computational cost. For the application of NeRF to large-scale scenes, a method based on NeRF is proposed in this paper to effectively use height data which can be obtained from GIS (Geographic Information System). For this purpose, the scene space was divided into multiple objects and a background using the height data to represent them with separate neural networks. In addition, an adaptive sampling method is also proposed by using the height data. As a result, the accuracy of image rendering was improved with faster training speed.

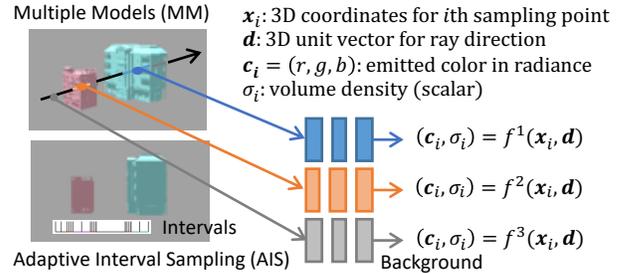

Fig. 1 Schematic of proposed method consisting of two techniques: Multiple Models (MM) and Adaptive Interval Sampling (AIS).

## 1. INTRODUCTION

Neural Radiance Fields (NeRF) [1] is a method for synthesizing a novel-view image of an object from multiple-view images by learning a volume rendering model represented by neural networks. Recently, a great progress has been made for this model, such as scalability of the target scenes [2, 3] and computational efficiency both in training and inference [4, 5]. It is expected to apply NeRF in wide range of applications, such as virtual reality and augmented reality.

Although most studies related to NeRF has focused on small objects, some researches have tried to extend the method to apply it to large-scale outdoor scenes. For example, NeRF++ [6] has applied the NeRF method to the scenes including both foreground objects and backgrounds by dividing the scene space into the inner spherical volume including the objects and the outer volume including backgrounds. Furthermore, Mega-NeRF [2] and Block-NeRF [3] have constructed 3D environments from large-scale visual captures spanning buildings or even multiple city blocks. Urban Radiance Fields [7] has also performed 3D reconstruction from less controlled data captured by mapping platforms such as Street View, incorporating additional LiDAR data.

In line with these researches, a method to apply the NeRF model to scenes with height data is proposed in this paper. Specifically, the proposed method is supposed to utilize GIS (Geographic Information System [8]) data to obtain height data of buildings in a city. To make effective use of the height data in the NeRF model, two techniques were incorporated into the model, as shown in Fig. 1. First, the scene was divided into multiple objects such as buildings based on the height data, to represent them by separate radiance-field models. Since it is assumed that the height data is available from GIS data, more accurate division is possible than the previous models such as NeRF++ [6] and Mega-NeRF [2]. Second, sampling points in the volume rendering of NeRF were adaptively set based on the height data. Since the borders of objects are more important than inside or outside of the objects, the sampling points were densely set at the borders. The proposed model was tested with scenes including multiple objects with height data.

The contributions of this paper include:
(1) A method of volume rendering model is proposed for the scenes with height data, which is assumed to be obtained from GIS data.
(2) To make effective use of the height data, the scene space is divided into multiple objects and a background based on the heights of the objects, to apply multiple rendering models for these objects and to set sampling points densely at the border of the objects.
(3) The proposed method produced better rendered images both qualitatively and quantitatively than previous methods.

## 2. RELATED WORKS

This section briefly reviews several solutions based on NeRF [1] for large-scale scenes and effective sampling methods.

DeRF [9] decomposes a scene into multiple cells by spatial Voronoi decomposition, to represent them independently using a small multi-layer perceptron (MLP). KiloNeRF [10] has achieved real-time rendering of NeRF by using thousands of tiny MLPs instead of one single large MLP. Each individual MLP only needs to represent parts of the scene, thus smaller and faster MLPs can be used. This method has accelerated rendering by three orders of magnitude compared to the original NeRF model.

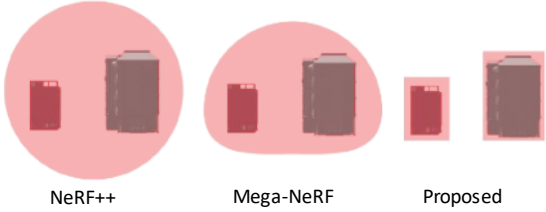

Fig. 2 Comparison among divided regions of scene space by proposed method and previous methods (NeRF++ [6] and Mega-NeRF [2]). The proposed method can more accurately divide objects such as buildings from the background using height data. Note that Mega-NeRF further divides the regions into grid cells irrespective to objects.

While NeRF and its related works have focused on small objects, NeRF++ [6] has been proposed to apply the NeRF model to outdoor scenes. Since the outdoor scenes include both objects and backgrounds far away from the objects, it has been difficult for NeRF to model the details of the objects with coarse background representations, simultaneously. Therefore, the scene space is divided into two separate volumes: the inner spherical volume containing all foreground objects, and the outer volume containing the remainder of the environment representing the backgrounds. These two volumes are separately represented by independent models. Furthermore, Mega-NeRF [2] has decomposed a scene into a set of spatial cells, learning a separate NeRF model for each only using the pixels whose rays intersect the spatial cell. Block-NeRF [3] has extended NeRF to render city-scale scenes spanning multiple blocks by decomposing the scene into blocks individually represented by NeRF models. This decomposition decouples rendering time from scene size, enables rendering to scale to arbitrarily large environments, and allows per-block updates of the environment. Urban Radiance Fields [7] has performed 3D reconstruction and novel view synthesis from less controlled data captured by mapping platforms such as Street View. To realize it, three components are incorporated: lidar information in addition to RGB signals, a separate model for camera rays pointing at the sky, and affine color transformation for each camera to compensate for varying exposure. Sat-NeRF [11] has been also proposed to learn a 3D reconstruction model from multi-view satellite photogrammetry in the wild by the end-to-end training.

The original NeRF has proposed the hierarchical volume sampling to efficiently render objects, where two networks for coarse and fine samplings are separately modeled. First, a model is constructed with coarse sampling, and then more informed sampling points are produced by calculating the weights representing contribution of each coarse sampling point to the rendering. The second model is then trained with the fine sampling refined from the coarse sampling. To improve the sampling strategy further, EfficientNeRF [4] has proposed valid and pivotal sampling at the coarse and fine stage, respectively, by analyzing the density and weight distribution of the sampling points.

In this paper, a method of NeRF is proposed for utilizing height data supposed to be obtained from GIS data, which has not been sufficiently explored so far.

## 3. METHOD

The proposed method divided the scene space into multiple objects and a background, as shown in Fig. 1, using height data supposed to be obtained from GIS data. The multiple objects and the background were modeled by multiple neural networks to represent neural radiance fields more accurately depending on the objects or the background. Representing the divided regions as separate models is the same idea as the previous works such as NeRF++ [6] and Mega-NeRF [2]. However, the proposed method is different from them in terms that more accurate division is possible since it is based on height data of objects, making more accurate modeling possible, as shown in Fig. 2. The reason to build separate models for different objects is that each object may have different characteristics, so that these specific characteristics are modeled by separate neural networks.

Furthermore, sampling points for volume rendering can be effectively set in the proposed method using the height data, since the borders of the objects are important for the volume rendering so that sampling points should be densely placed at the borders.

In the following subsections, the original NeRF model is first explained, and then the two components of the proposed method, multiple models (MM) and adaptive interval sampling (AIS), are explained in detail.

### 3.1. Neural Radiance Fields (NeRF)

NeRF [1] is a method for generating free viewpoint images from multiple photos taken from various angles. In the NeRF model, a static scene is represented as a continuous 5D function that outputs the radiance emitted in each direction $(\theta, \phi)$ at each point $(x, y, z)$ in the scene space, and a density at each point which acts like a differential opacity controlling how much radiance is accumulated by a ray passing through the point $(x, y, z)$. When the direction $(\theta, \phi)$ is expressed in a 3D Cartesian unit vector $d$, each 3D coordinate $(x, y, z)$ as $x$, the emitted color in the radiance as $c = (r, g, b)$, and the volume density as $\sigma$, the function representing the radiance field is expressed by

$$(c, \sigma) = f(x, d), \qquad (1)$$

where a multi-layer perceptron (MLP) is used with positional encoding for representing the function.

An image is rendered using volume rendering from the neural radiance field in Eq. 1 represented by the volume density $\sigma$ and directional emitted radiance $c$ at each point in the scene space. The discrete form of the volume rendering [1] can be expressed by

$$\hat{C}(r) = \sum_{i=1}^{N} T_i(1 - \exp(\sigma_i \delta_i))c_i \qquad (2)$$

$$T_i = \exp\left(-\sum_{j=1}^{i-1} \sigma_j \delta_j\right) \qquad (3)$$

where $r$ is a ray in 3D space corresponding to a pixel in the rendered image, $\hat{C}(r)$ is pixel values $(r, g, b)$ in the rendered image, $N$ is the number of sampling points along the ray, $i$ represents $i$-th sampling point, and $\delta_i$ is the $i$-th sampling interval. A sampling point is drawn from a uniform distribution in each sampling interval to train the network at continuous positions as much as possible. The loss function for training the network is the mean squared error between an image given in the training dataset and the rendered image with same camera parameter.

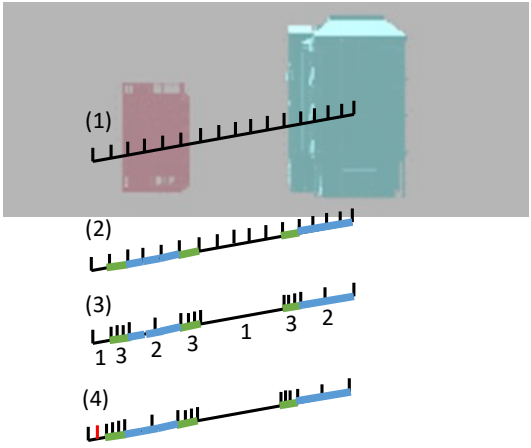

Fig. 3 Procedure for setting sampling intervals along ray in Adaptive Interval Sampling (AIS).

### 3.2. Multiple Models (MM)

The proposed method divided the scene space into multiple objects such as buildings and a background using height data of the objects to express multiple objects with multiple MLP models. By representing a smaller region of the scene with a model, the expressive power is increased so that more accurate rendering can be expected.

To divide MLP models for objects based on the height data, 3D coordinates $(x, y, z)$ in the 3D space were divided into $M$ groups representing $(M-1)$ object groups and a background group. To do this, the 2D coordinates $(x, y)$ were first divided into two groups based on the height data (object height, not $z$) at the coordinate $(x, y)$. The 2D coordinates with the height larger than a threshold were divided into the object group, and the rest was divided into the background group, since the height data would be close to zero at the points where objects do not exist. Then, the 2D coordinates $(x, y)$ in the object group were divided into $(M-1)$ groups using k-means clustering based on the 2D coordinates. Thus, neighbouring 2D coordinates were grouped together to produce a group for each object or neighbouring objects. For dividing 3D points in the 3D space, each point $(x, y, z)$ at the 2D coordinate $(x, y)$ in the $m$-th object group was divided into the $m$-th 3D object group if $z \leq$ height, and into the 3D background group if $z >$ height, since the 3D points $(x, y, z)$ above the object (height) would be the background. For the points in the 3D space grouped into the background in the 2D coordinate, all points were divided into the 3D background group in the 3D space, since there would be no object at the point. Thus, the 3D scene space was divided into $M$ regions ($(M-1)$ groups for objects and a background group) depending on the height data. In the proposed method, each region was modeled by an independent MLP model, represented by

$$(c, \sigma) = f^m(x, d), \quad (4)$$
$$x \in S_m \text{ (Set of 3D cordinates in } m\text{th group)}$$

### 3.3. Adaptive Interval Sampling (AIS)

As explained in the related works, the original NeRF has proposed the hierarchical volume sampling to efficiently render objects, where a model is constructed with coarse sampling and then more informed sampling points are produced so that a fine model can be trained with the fine sampling.

Since it is assumed that the height data is available in the proposed method, more important sampling points can be set in the coarse sampling stage. The sampling points at borders of objects would be more important than the inside or outside of the objects. Therefore, the sampling interval was set based on height data of objects with the following procedure (also shown in Fig. 3).

(1) $N$ sampling intervals are evenly set along a ray.
(2) Each sampling interval is classified into Background, Border, or Object based on the height data. If $z >$ height for both ends of the interval, the interval is classified into Background ('height' is the object height, whereas $z$ is taken from 3D coordinates $(x, y, z)$ at the ends of interval). If $z \leq$ height for both ends of the interval, the interval is classified into Object. Otherwise, the interval is classified into Border. Object and Border are represented in blue and green lines in Fig. 3, respectively.
(3) After neibouring intervals are integrated if the labels are the same, it is divided into 3 and 2 intervals for Border and Object, respectively. Since the border is important for the rendering, a short sampling interval is assigned for dense sampling. Note that the values of 3 and 2 are hyper parameters, and were not optimized.
(4) If the number of sampling intervals is less than $N$, the closest interval to the camera is divided to obtain $N$ sampling intervals. On the other hand, if it is more than $N$ intervals, the most far intervals were integrated to be $N$ sampling interval, since the most far points would not be important for the rendering in this case.

Following the original NeRF, a point was sampled from a uniform distribution for each sampling interval obtained by above procedure. This strategy of sampling is called Adaptive Interval Sampling (AIS) in this paper. After constructing a coarse model based on the coarse sampling with AIS, the fine sampling stage was applied in the same way as the original NeRF method.

## 4. EXPERIMENTS

### 4.1. Experimental Setup

Two synthetic datasets were prepared for evaluating the proposed method. The first dataset was images of Legos, whose model was obtained from DeepVoxels [12]. As shown in Fig. 4(a), the same 9 Lego models were placed side by side to simulate a scene where multiple objects were placed together. For this dataset, 100 images were taken in 256×256 pixels with various camera parameters, where two situations were tested: the datasets without and with images taken from above. The test without images from above was conducted for simulating the situation when the available images were limited.

The second dataset was images of buildings in a city model [13], as shown in Fig. 4(b). The area of the model was 1200m×1200m, with height data in the grid of 5m×5m. For this dataset, 300 images were taken in 512×512 pixels.

The architecture of MLP, the learning rate and other parameters were set to the same as in NeRF [1]. The ray sampling intervals was set to $N = 64$. The number of models in MM was set to $M = 10$ for the Legos dataset and $M = 6$ for the buildings dataset which was limited by the GPU memory size. The models were trained for 200,000 epochs on a single GPU (Nvidia RTX3090). The accuracy of the methods was evaluated by PSNR (Peak Signal-to-Noise Ratio) [14].

### 4.2. Results

The evaluation results of the proposed methods are shown in Table 1, compared with the conventional methods (NeRF [1] and NeRF++ [6]). For all the datasets, the proposed method with MM and AIS achieved the highest score of PSNR, indicating that the height data was effective for accurate rendering based on NeRF. Furthermore, it can be seen from the table that both MM and AIS contributed to the accurate image rendering. The learning curves shown in Fig. 5 proved that the training speed in the proposed method was faster than the original NeRF. The accuracy in NeRF at 50,000 epochs was obtained at 10,000 epochs in the proposed method.

The qualitative comparison is shown in Fig. 6. This is an example of rendered images on the Legos dataset with images from above. It is clear that the proposed method impressively improved the quality of the image by both MM and AIS. On the other hand, it was still challenging that the expressive capability declined as the distance from the center increased. Since this is a common problem in NeRF, further studies would be required to improve the quality of image rendering.

### 5. CONCLUSION

In this paper, a method of volume rendering was proposed for the situation when the height data of objects can be obtained from GIS data. The scene space was divided into multiple objects and a background, each of which is separately modeled by neural networks. The algorithm for accurately dividing the space was designed using the height data. In addition, adaptive interval sampling was proposed to set sampling points at the borders of the objects based on the height data. It can be seen from the evaluation results that the proposed method outperformed the baseline methods both qualitatively and quantitatively. Furthermore, the multiple models and the adaptive interval sampling were both effective for improving the accuracy of image rendering. It was confirmed that the training speed was also improved by the proposed method.

Although the accuracy was improved even for the situation with limited number of images (such as Legos without images from above) by the proposed method, the rendered image taken from above was still challenging. One possible solution would be to use satellite images during the training of the NeRF model combined with the height data from GIS. This may be a solution to reconstruct a whole city from the limited number of images.

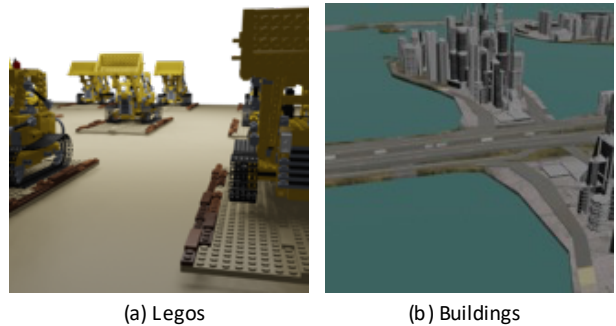

(a) Legos　　　　　　(b) Buildings

Fig. 4 Sample images in training datasets.

Table 1 Quantitative comparison with baselines in PSNR [dB].

| | Methods | Legos w/o images from above | Legos w/ images from above | Buildings |
|---|---|---|---|---|
| Conventional | NeRF [1] | 27.130 | 30.044 | 28.327 |
| | NeRF++ [6] | 29.003 | 31.520 | 28.032 |
| Proposed | MM | 28.881 | 31.858 | 29.910 |
| | AIS | 29.123 | 31.040 | 29.010 |
| | MM+AIS | **30.816** | **32.313** | **30.024** |

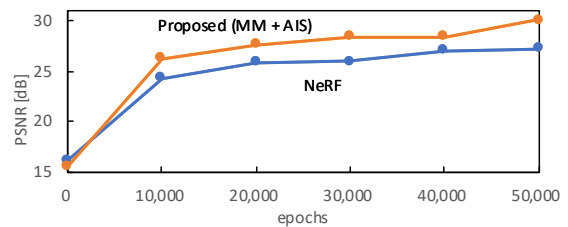

Fig. 5 Learning curves in Legos dataset without images from above on single GPU.

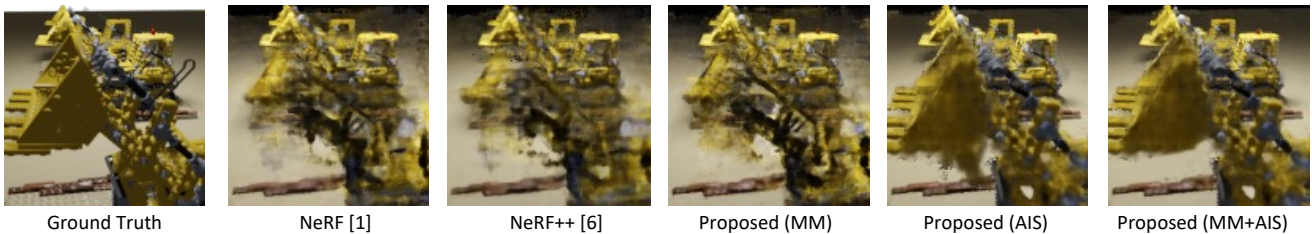

Ground Truth　　NeRF [1]　　NeRF++ [6]　　Proposed (MM)　　Proposed (AIS)　　Proposed (MM+AIS)

Fig. 6 Qualitative comparison with other methods in Legos dataset with images from above.